\documentclass[10pt,twocolumn,letterpaper]{article}

\usepackage[pagenumbers]{cvpr}      

\definecolor{cvprblue}{rgb}{0.21,0.49,0.74}
\usepackage[pagebackref,breaklinks,colorlinks,allcolors=cvprblue]{hyperref}
\usepackage{multirow}

\def\paperID{*****} 
\def\confName{CVPR}
\def\confYear{2025}

\title{Semantic Compression of 3D Objects for Open and Collaborative Virtual Worlds}

\author{
Jordan Dotzel\textsuperscript{*} \quad
Tony Montes\textsuperscript{*} \quad
Mohamed S. Abdelfattah \quad
Zhiru Zhang \\
{\tt\small \{dotzel,tsm85,mohamed,zhiruz\}@cornell.edu} \\
School of ECE, Cornell University
}

\begin{document}
\maketitle
\renewcommand{\thefootnote}{\fnsymbol{footnote}}
\footnotetext[1]{Equal contribution}

\begin{abstract}
Traditional methods for 3D object compression operate only on structural information within the object vertices, polygons, and textures.
These methods are effective at compression rates up to $10\times$ for standard object sizes but quickly deteriorate at higher compression rates with texture artifacts, low-polygon counts, and mesh gaps.
In contrast, semantic compression ignores structural information and operates directly on the core concepts to push to extreme levels of compression.
In addition, it uses natural language as its storage format, which makes it natively human-readable and a natural fit for emerging applications built around large-scale, collaborative projects within augmented and virtual reality.
It deprioritizes structural information like location, size, and orientation and predicts the missing information with state-of-the-art deep generative models.
In this work, we construct a pipeline for 3D semantic compression from public generative models and explore the quality-compression frontier for 3D object compression.
We apply this pipeline to achieve rates as high as $10^5\times$ for 3D objects taken from the Objaverse dataset and show that semantic compression can outperform traditional methods in the important quality-preserving region around $100\times$ compression.
\end{abstract}   
\section{Introduction}
\label{sec:intro}

\begin{figure} [t]
    \centering
    \includegraphics[width=1\linewidth]{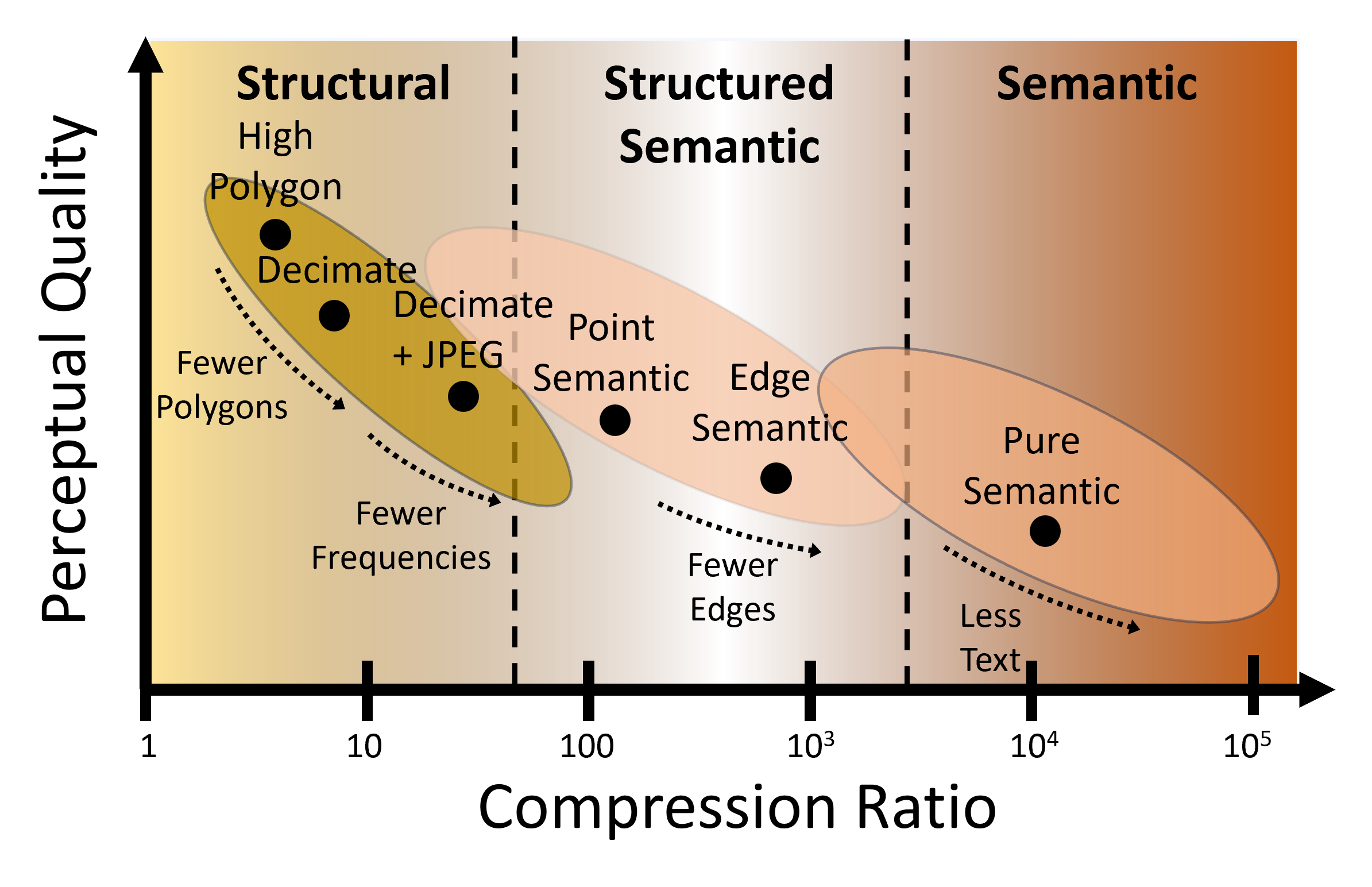}
    \vspace{-10pt}
    \caption{\textbf{Semantic Compression:} Extreme compression requires primarily storing semantic as opposed to structural information, such as point clouds, polygons, or frequencies. Semantic compression enables extreme ratios by preserving human-oriented semantics as opposed to structural information or its derivatives. In particular, the structured semantic compression outperforms traditional methods in the region around $10-100\times$ compression, beyond which only semantic-based methods can be applied.}
    \label{fig:semantic}
\end{figure}

The growing popularity of augmented and virtual reality platforms are set to enable near limitless virtual worlds where users can manipulate an ever-increasing number of objects.
To make this possible, devices will need large resources to store textures, 3D assets, and animations that must be transferred over the network and stored locally on device.
This requires new methods for compression and decompression algorithms that can handle the growing amount of required storage.
Semantic compression has been proposed as a new paradigm for storing extremely compressed data~\cite{weissman2023textual}.
It focuses on preserving only the human-centric, \textit{semantic} information in the input by disregarding precise, structural information.  
These semantics are typically represented with natural language descriptions, which have co-evolved with human society to efficiently describe the most important concepts.
These descriptions lead to extremely small representations at the cost of lost structural information such as pixels or polygons.

The quality-compression tradeoffs are summarized in Figure~\ref{fig:semantic}, which divides the compression landscape into structural, semantic, and structured semantic regions.
Most traditional methods are found within the structural region, such as JPEG for images, MPEG for videos, and decimated meshes for 3D objects.
This work, on the other hand, introduces methods for structured semantic compression using edge maps and pure semantic compression using natural language.
It further claims that optimal compression with respect to human preferences must transition between the structural and semantic regions, and that this central region outperforms traditional structured compression for most objects.

Overall, semantic compression gives rise to two primary advantages over traditional compression: orders of magnitude more objects and storage natively within a human-readable format.
This opens up a new set of applications, from augmented and virtual reality, to direct and private storage within brain-computer interfaces.
It enables shared virtual worlds with millions of objects that can be collaboratively edited by non-experts.
Since the storage format is human-readable, it is portable across many systems and provides the most general interface to end-users.

\begin{figure} [t]
    \centering
    \includegraphics[width=1\linewidth]{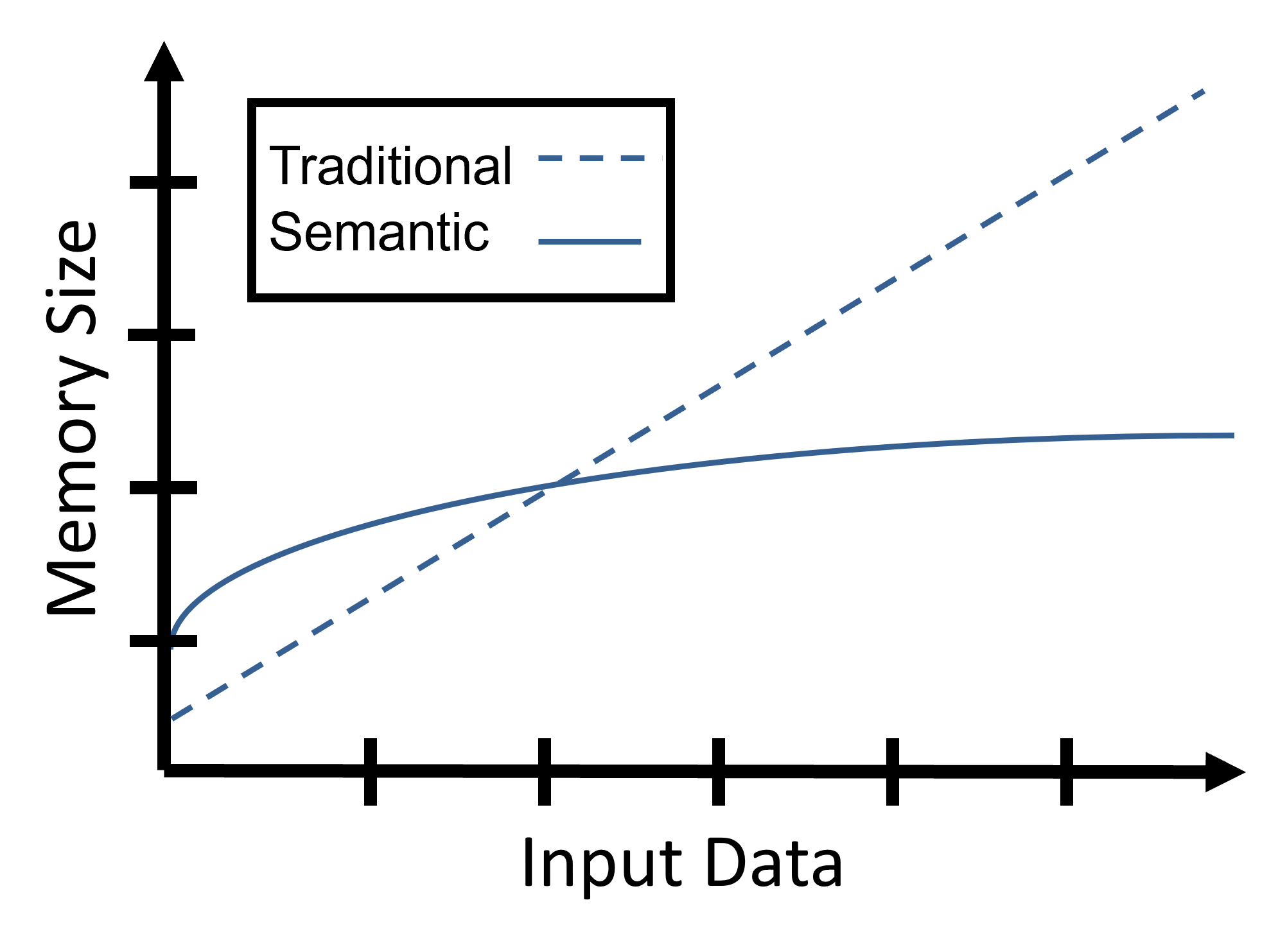}
    \vspace{-10pt}
    \caption{\textbf{Semantic Scaling:} As the size of the input grows, with more objects and more complex objects (higher resolution, denser point clouds, or more polygons), semantic compression becomes more efficient. Traditional compression scales linearly with the input, while semantic compression scales sub-linearly with the semantic content. The initial memory for the compressed world model is amortized over many objects.}
    \label{fig:memory}
\end{figure}

Another key advantage is its scalability with input data size, as shown in Figure~\ref{fig:memory}.
Traditional methods scale linearly with the input structural data, like resolution or polygon count, while semantic compression scales with the semantic complexity, such as the number of independent concepts in the input.
With most data, the semantic complexity is sub-linear in the input size since most of the structural resolution is redundant and can be regenerated with trained models.
It also has privacy implications since certain sensitive features can be removed or obfuscated in storage, such as facial features or other personally identifying information.

Despite these advantages, semantic compression comes with additional challenges, especially related to increased computations.
It shifts the burden from memory and storage to additional computations that are required to regenerate 3D objects during decompression.
With current technology, this requires large multi-modal language models that require billions of floating-point operations.
Furthermore, the language model itself can consume a lot of memory, therefore, semantic compression must be used with tens or hundreds of objects to amortize this cost over a large number of stored objects.
This, however, fits well with emerging applications in the metaverse which contain vast numbers of objects.

In this work, we explore the quality-compression tradeoffs for traditional, structured semantic, and semantic compression of 3D objects, discuss the high-level advantages of semantic compression, and highlight the current technological limitations to guide future work.
We summarize our contributions as follows:
\begin{enumerate}
    \item Construct the first semantic compression pipeline for 3D objects that outperforms traditional methods in the region beyond $10\times$ compression in terms of human-evaluated quality and compression rate.
    \item Explore the quality-performance trade-offs of extreme semantic compression with and without structural information for 3D objects.
    \item Compare automated metrics with human evaluations for 3D reconstructions and highlight the limitations of F-Score and CLIP with semantic compression.
\end{enumerate}

\section{Background}
\label{sec:background}

\begin{figure} [t]
    \centering
    \includegraphics[width=1\linewidth]{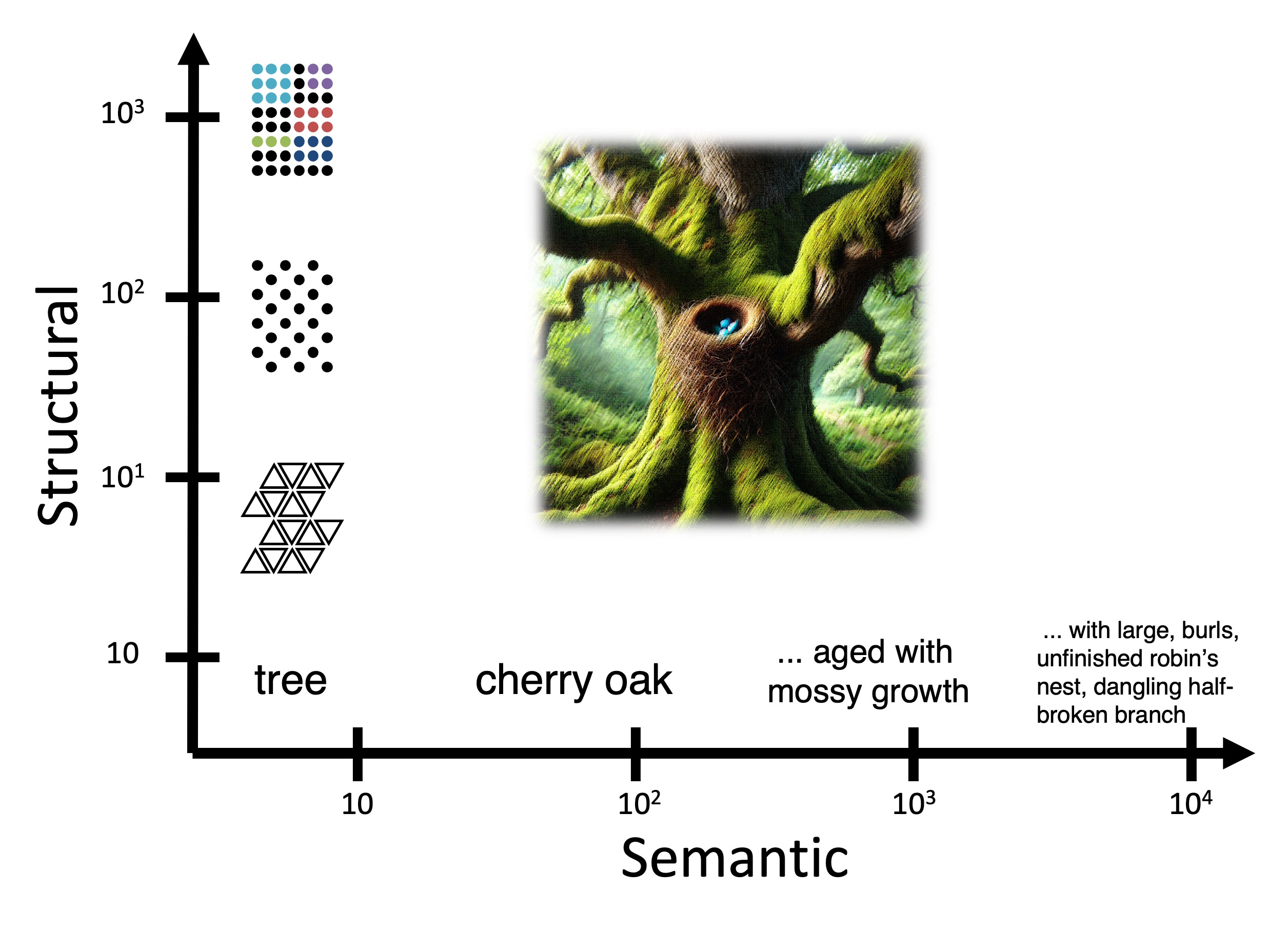}
    \vspace{-10pt}
    \caption{\textbf{Semantic Factorization:} Across most media formats, the structural and semantic dimensions can be factorized and compressed separately. This stores 3D point clouds or 2D projections in addition to natural language description.}
    \label{fig:factorization}
\end{figure}

Traditional compression methods have achieved impressive compression rates by removing human-imperceptible information.
These \textit{perceptual compression} methods, however, are limited by operating within the structural input space (e.g., pixels or voxels) or direct transformations of them (frequencies or polygons).
JPEG, for instance, operates within the frequency domain and removes the least perceptible frequencies according to empirical quantization tables.
Semantic compression instead functions within the semantic domain and maps out directions that minimize the perceptual differences.

It can also be combined with light-weight structural hints such as sparse point clouds or 2D edge maps to form a hybrid structured semantic approach. This approach attempts to operate on the factorized space of semantic and structural information and compress them separately, as shown in Figure~\ref{fig:factorization} for 3D objects.
The structural information, as measured in bits, increases from polygons, to point clouds, to colored voxels, while the semantic information progressively includes more detailed concepts.
These two dimensions, while not being completely orthogonal, align well with human preferences, preferring structural fidelity for low compression rates and semantic quality at high compression rates.

\subsection{Generative Models}
\label{sec:generative}

\begin{figure} [t]
    \centering
    \includegraphics[width=1\linewidth]{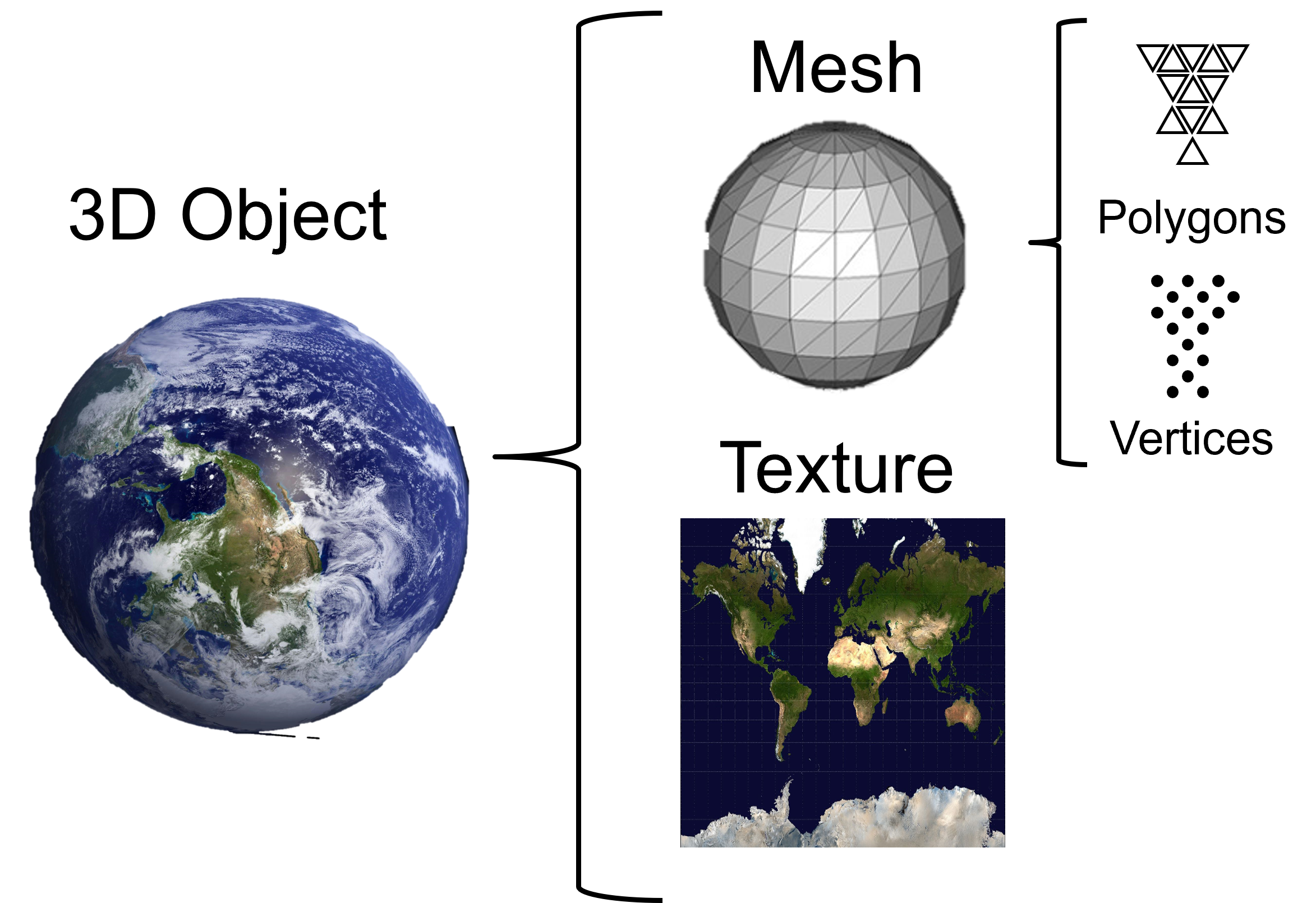}
    \vspace{-10pt}
    \caption{\textbf{3D Objects:} 3D objects are composed of files for the texture and mesh, where the mesh breaks down into vertices and polygons between them. The mesh typically dominates the size of the object.}
    \label{fig:3d_format}
\end{figure}

Until the last few years, it was not possible to accurately map out the semantic space in an automated way.
However, recent advances in generative models, especially transformer-based diffusion models~\cite{rombach2022stable, zhang2023adding, bhown2018humans} and large language models~\cite{gemini2024gemini, openai2023gpt4, groeneveld2024olmo}, have enabled semantic compression in a growing number of fields.
These models can transform between the input space and natural language, prioritize semantic features over others, condition on structural information, and then transform back into the original space.
Unlike discriminative models, which learn the function \( p(y | x) \) and predict a label \( y \) for a given observation \( x \), generative models learn \( p(x) \) or \( p(x, y) \).
This means that generative models directly approximate the input distribution, and then semantic compression can use this compressed distribution to avoid storing redundant structural detail.

This work focuses on diffusion models, such as DALLE3 from OpenAI, which have become the standard generative architecture for visual tasks.
They operate by transforming a data sample through iterative noise addition and removal. The process consists of a \textit{forward diffusion} phase, where noise is incrementally added to the data, and a \textit{reverse process}, where the model learns to remove this noise step-by-step, reconstructing a sample from noise. The forward diffusion process is governed by:
\[
q(x_t | x_{t-1}) = \mathcal{N}(x_t; \sqrt{1 - \beta_t} x_{t-1}, \beta_t I),
\]
where \( \beta_t \) is a noise schedule parameter that determines the level of noise added at each step \( t \).
In addition, diffusion models like ControlNet~\cite{zhang2023adding} enable conditional generation by introducing guidance variables (e.g. additional structural information like edges).
When used in semantic compression, this allows storing additional structural information to be used during decompression.

\subsection{3D Objects}
\label{sec:3d_models}

\begin{figure} [t]
    \centering
    \includegraphics[width=1\linewidth]{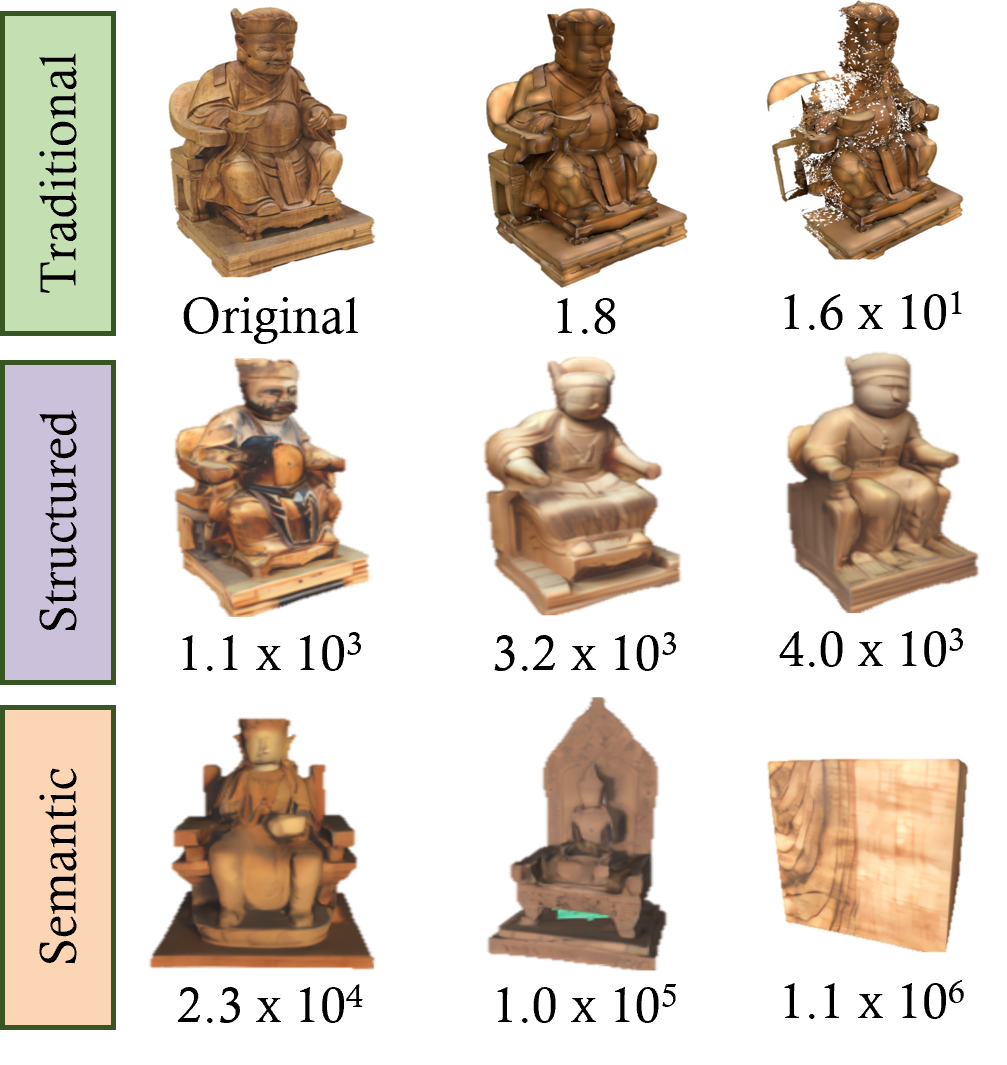}
    \vspace{-10pt}
    \caption{\textbf{Semantic Compression}: Further semantic compression loses more and more detail with the benefit of significant memory savings. Top row includes the original with additional structural information, and the bottom row is pure semantic compression. Compression ratios are listed under each object.}
    \label{fig:progression}
\end{figure}

Recent advances in 3D diffusion models~\cite{liu2023one2345, shi2023zero123plus, Qian2023Magic123OI, Long2023Wonder3DSI} and the availability of large-scale 3D object datasets like Objaverse~
\cite{deitke2022objaverse, deitke2023objaversexl} have opened up possibilities of semantic compression of 3D scenery.
These objects contain a combination of mesh information, including vertices, polygons, normal vectors, and texture information, as shown in Figure~\ref{fig:3d_format}.
Traditional compression of 3D objects relies on the separate traditional compression of the mesh and texture.
For the mesh, methods like decimation are popular, which iteratively merges vertices and polygons from the mesh~\cite{garland2023mesh} and can typically achieve compression ratios up $10\times$ depending on the input (Appendix~\ref{sec:decimate} for more detail).
For the texture, traditional image compression using JPEG is the most common, yet since the mesh typically dominates the file size, texture compression has more limited benefit.
These techniques combined together typically can only achieve rates up to $15\times$ without significant quality degradation, which opens opportunities for semantic compression to improve further.

\section{Method}
\label{sec:method}

\begin{figure} [t]
    \centering
    \includegraphics[width=1\linewidth]{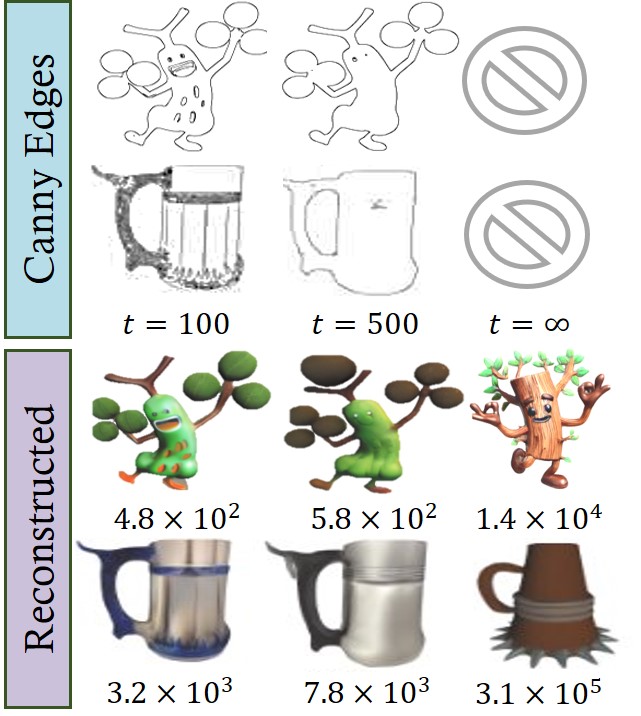}
    \vspace{-5pt}
    \caption{\textbf{Structural Control:} Higher thresholds $t$ lead to less detail in the edge map and therefore higher compression rates, yet this comes at the cost of lost fine-grained detail. The compression rates are listed underneath each object and increase with less structural control.}
    \label{fig:structure}
\end{figure}

\begin{figure*} [t]
    \centering
    \includegraphics[width=1\linewidth]{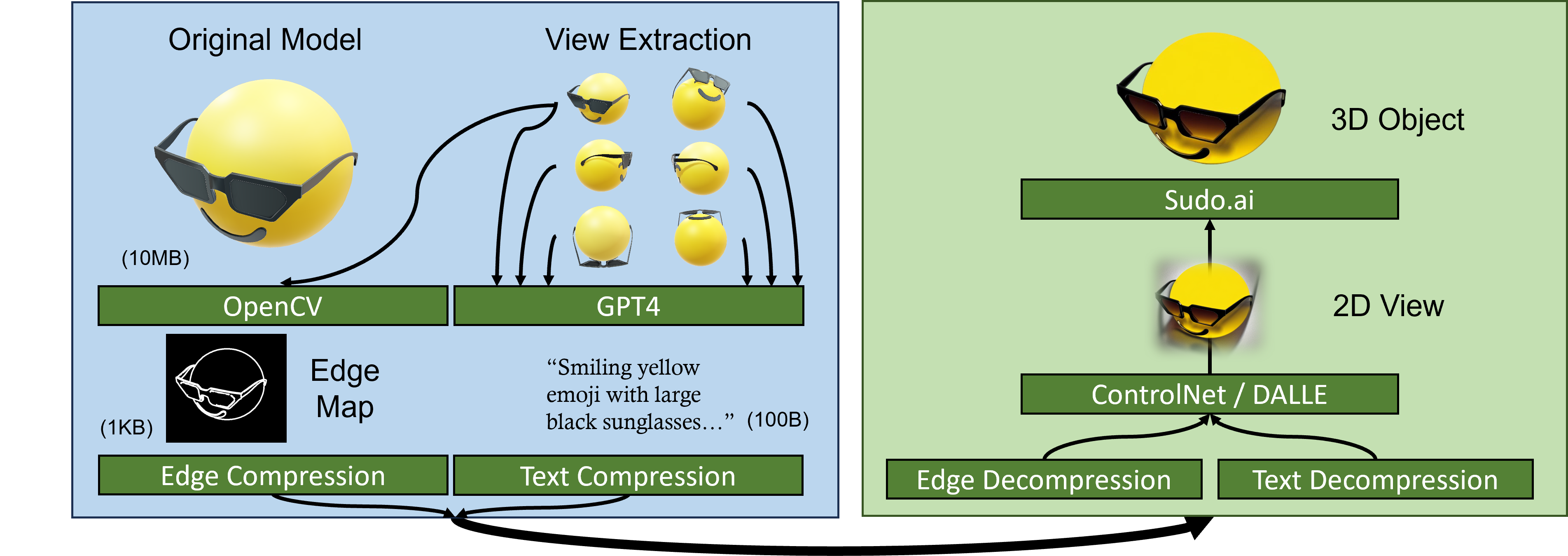}
    \vspace{-10pt}
    \caption{\textbf{Method Overview:} Multiple views are extracted at fixed orientations from the object, which are then passed into GPT4 for structured descriptions. This description is then further compressed and optionally combined with a compressed edge map from the frontal view. These data are then decompressed and passed to the (conditioned) image generative model, either ControlNet or DALLE3. This view is finally passed to sudo.ai to project from one view to a corresponding mesh and texture files.}
    \label{fig:method}
\end{figure*}

This work builds a semantic compression pipeline using public models with optional structural control using edge maps.
The inputs and outputs of this process are shown in Figure~\ref{fig:progression}.
The original wooden statue is passed to the pipeline to be compressed using structured semantic or pure semantic compression.
This leads to relatively high compression ratios compared to the mesh- and texture-based methods shown in the first row.
The entire pipeline is shown in detail in Figure~\ref{fig:method}.

\subsection{Compression}
\label{sec:compression}

\begin{table}[t]
\centering
\footnotesize
\setlength{\tabcolsep}{5pt}  
\begin{tabular}{ccccccc}
\toprule
& \multicolumn{4}{c}{\textbf{Threshold}} & \\
\cmidrule(lr){2-5}
\textbf{Resolution} & \textbf{50} & \textbf{300} & \textbf{550} & \textbf{800} & \textbf{Breakeven} \\
\midrule
64   & 83.7$_{\text{\scriptsize{3.9}}}$ & 92.2$_{\text{\scriptsize{2.6}}}$ & 93.0$_{\text{\scriptsize{3.2}}}$ & 93.3$_{\text{\scriptsize{3.1}}}$ & 87.98 \\
128  & 86.4$_{\text{\scriptsize{5.9}}}$ & 95.5$_{\text{\scriptsize{1.7}}}$ & 96.3$_{\text{\scriptsize{2.0}}}$ & 96.3$_{\text{\scriptsize{2.0}}}$ & 89.70 \\
256  & 89.1$_{\text{\scriptsize{7.4}}}$ & 97.4$_{\text{\scriptsize{0.6}}}$ & 98.0$_{\text{\scriptsize{1.0}}}$ & 98.1$_{\text{\scriptsize{1.1}}}$ & 90.98 \\
512  & 92.0$_{\text{\scriptsize{7.0}}}$ & 98.5$_{\text{\scriptsize{0.1}}}$ & 99.0$_{\text{\scriptsize{0.5}}}$ & 99.0$_{\text{\scriptsize{0.6}}}$ & 91.99 \\
1024 & 95.4$_{\text{\scriptsize{4.5}}}$ & 99.3$_{\text{\scriptsize{0.1}}}$ & 99.5$_{\text{\scriptsize{0.3}}}$ & 99.5$_{\text{\scriptsize{0.3}}}$ & 92.79 \\
2028 & 98.0$_{\text{\scriptsize{1.9}}}$ & 99.6$_{\text{\scriptsize{0.1}}}$ & 99.7$_{\text{\scriptsize{0.1}}}$ & 99.7$_{\text{\scriptsize{0.2}}}$ & 93.43 \\
\bottomrule
\end{tabular}
\caption{\textbf{Edge Map Sparsity:} Nearly all object views have sufficient sparsity to benefit from the COO format. The breakeven sparsity is the sparsity necessary for the COO format outperform the dense format. Values are averaged across 100 views of objects.}
\label{tab:sparsity_table}
\end{table}

The first step of the compression pipeline involves extracting six views of the input object at fixed orientations.
These views are then concatenated and passed to the OpenAI GPT4 model~\cite{openai2023gpt4} for a descriptive summary (prompts listed in Appendix~\ref{sec:prompts}).
The additional views capture information from all sides of the input image that could be potentially lost with just a single forward or top view.
This description is then further compressed in another request to limit it to the $d$ most important characters.
Since this is a soft request to GPT4, the result is filtered and clamped to ensure the representation does not exceed $d$ characters, including spaces, and only includes lowercase English letters.
Any language in theory can be used for semantic compression, but the English models are the most mature.

For structural semantic compression, the frontal view of the object is passed to OpenCV to extract an edge map to preserve the approximate structure of the view.
The level of detail in the output map is controlled through a joint threshold of the pixel gradient magnitudes.
Higher thresholds reduce the edge detail by approximately removing all gradient magnitudes less than this threshold.
This edge map is returned as a dense binary image, where each pixel is either black or white, and then it is further compressed by using the sparse coordinate format (COO) and downsampling so each coordinate can be represented with an 8-bit integer.
Profiling in Table~\ref{tab:sparsity_table} shows that most natural views of objects are sparse enough across resolutions that they benefit from sparse storage.

\subsection{Structural Control}
\label{sec:strcutural}

At the core of this structural semantic compression method is the Canny edge detection algorithm~\cite{canny1986canny}.
This begins with smoothing the image with a Gaussian filter and then calculating its gradients to identify edges.
The gradient of the image \( I_G \) is computed in the \( x \)- and \( y \)-directions as \( I_x \) and \( I_y \), respectively.
The gradient magnitude \( M(x, y) \) and orientation \( \theta(x, y) \) are given by:

\[
M(x, y) = \sqrt{I_x^2 + I_y^2}
\]
\[
\theta(x, y) = \arctan\left(\frac{I_y}{I_x}\right)
\]

Pixels are then classified as edges based on two thresholds: \( t_{high} \) and \( t_{low} \).
Pixels with gradient magnitudes \( M(x, y) \) above \( t_{high} \) are considered \textit{strong edges}. Those with \( M(x, y) \) between \( t_{low} \) and \( t_{high} \) are classified as \textit{weak edges} if they are connected to strong edges; otherwise, they are suppressed.

This yields the final edge map, where only weak edges and connected strong edges are preserved.
For simplicity, we fix the ratio of the two thresholds as:

\[
t = t_{high} = 2 \cdot t_{low}
\]

This ratio simplifies the hyper-parameters to only a single threshold, $t$, which then determines the level of structural detail.
By tuning this threshold, the edge map can focus on only significant contours and ignore the smaller detail in the image, which makes its overall structure relatively sparse.
This is shown in Figure~\ref{fig:structure} for increasing $t$ and therefore less edge detail.
The tree creature loses its facial features as $t$ increases from 100 to 500, and the cup gets a simpler design.
In the limit without any structure, the model attempts to capture the underlying ideas instead.

\subsection{Decompression}
\label{sec:decompression}

The decompression process takes the compression text and optional edge embeddings and attempts to recreate the original data.
First, it recreates the primary view based on the uncompressed description, routing either to DALLE3 in the unstructured case or ControlNet~\cite{zhang2023adding} with additional edge information.
DALLE3 offers high quality results across a large number of themes, yet it currently offers no structural conditioning.
ControlNet, on the other hand, is built from the Stable Diffusion model and has weaker overall performance but accepts multiple forms of structural control, including Canny edge, depth, and HED (holistically-nested edge detection) maps.
Within these options, the Canny edge maps offer the most efficient structural information and highest quality of results in practice.
The background from the returned view is then filtered away to isolate only the object for the next steps.

This filtered, reconstructed view is then passed to the Zero123++ image-to-3D generative model~\cite{shi2023zero123plus}, through its API access with sudo.ai.
This model uses this image to synthesize multiple views and then transforms them into a coherent 3D object, including mesh and texture files.
Detailed examples for each intermediate step are listed in the Appendix~\ref{sec:examples}.

\subsection{Comparison}
\label{sec:comparison}

\begin{figure*} [t]
    \centering
    \includegraphics[width=1\linewidth]{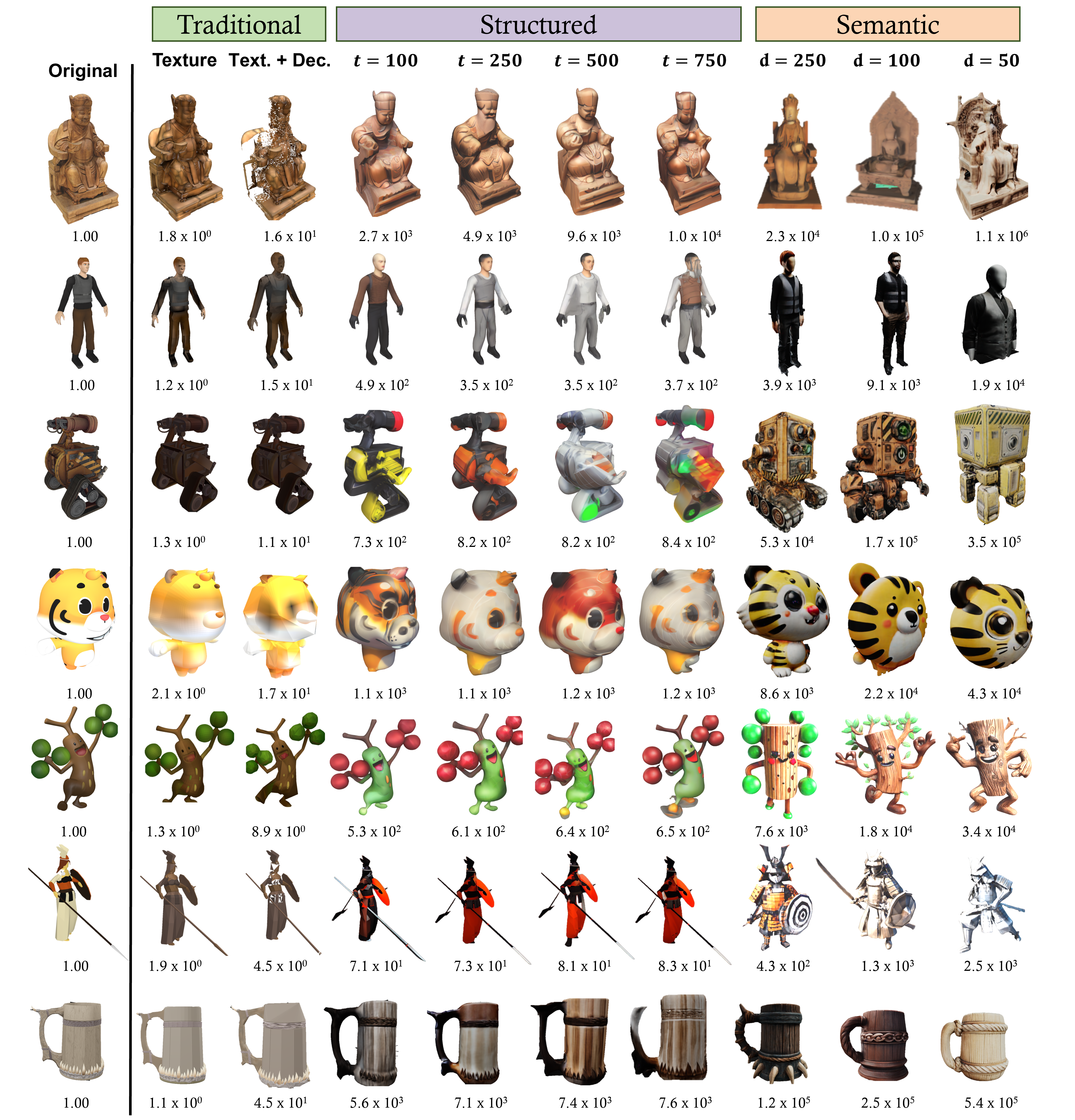}
    \vspace{-10pt}
    \caption{\textbf{Compression Examples:} Traditional compression is limited to within $15\times$ for most objects while maintaining quality, while structured semantic and pure semantic methods can push orders of magnitudes further. The edge threshold $t$ controls the amount of detail in the edge map, and the character count $d$ limits the semantic content.}
    \label{fig:examples}
\end{figure*}

A large-scale comparison across methods, hyper-parameters, and objects is shown in Figure~\ref{fig:examples}.
The compression rate increases from left to right, beginning with traditional mesh and texture compression, structured semantic compression, and then finally pure semantic compression.
The edge threshold $t$ is listed for the structured semantic compression with a fixed number of characters $d = 250$ for all methods.
Since the edge maps dominate the compressed file size, the number of characters is not too important, and the quality of the ControlNet outputs saturates quickly with description length so larger descriptions are not useful.

For pure texture compression, this figure shows that the object loses quality very quickly for relatively small compression rates.
Since most objects are dominated by the mesh size, texture compression can be fairly insignificant, but the exact compression rate depends on the original resolution and mesh-texture memory balance of the object.
Combined with the mesh decimation, objects begin to reach more useful compression rates.
Yet, they typically still cannot exceed $15\times$, depending on the original vertex and triangle count, without low-polygon artifacts and mesh gaps.

This highlights the usefulness of semantic compression, especially with some structural support.
In the figure, the higher the edge threshold, the less detail remains in the edge map, leading to higher compression and worse quality on average, although ControlNet has significant variance in its results for the same prompt (as shown in Appendix~\ref{sec:controlnet}).
More broadly, ControlNet has difficulty handling color (tree creature colors are inverted) and produces significant hallucinations.
For example, in Figure~\ref{fig:examples} tree creature arms are colored red likely because they resemble cherries, despite no mention of red in any of the descriptions.
These are short-term limitations of the current generation of models, not issues with the method itself.
Overall, structured semantic compression can scale more easily beyond $10\times$ compression compared to traditional compression, which succumbs early to mesh and texture artifacts.

In Figure~\ref{fig:examples}, the last three columns in each row represent pure semantic compression using the more powerful DALLE3 model.
For many objects, such as the wooden statue, the results are sufficiently strong to be used as replacement in most contexts.
Yet, other objects still fail with current neural networks but likely will succeed with stronger generative models.
\section{Evaluation}
\label{sec:evalatuion}

\begin{table*}[t]
\centering
\footnotesize
\setlength{\tabcolsep}{2.3pt}  
\begin{tabular}{lcccccccccccccccccccccccc}
\toprule
& \multicolumn{4}{c}{\textbf{Statue}} & \multicolumn{4}{c}{\textbf{Human}} & \multicolumn{4}{c}{\textbf{Robot}} & \multicolumn{4}{c}{\textbf{Tiger}} & \multicolumn{4}{c}{\textbf{Tree}} & \multicolumn{4}{c}{\textbf{Average}} \\
\cmidrule(lr){2-5} \cmidrule(lr){6-9} \cmidrule(lr){10-13} \cmidrule(lr){14-17} \cmidrule(lr){18-21} \cmidrule(lr){22-25}
\textbf{Method} & \textbf{F}$\uparrow$ & \textbf{C}$\uparrow$ & \textbf{MR}$\downarrow$ & $\times$ & \textbf{F}$\uparrow$ & \textbf{C}$\uparrow$ & \textbf{MR}$\downarrow$ & $\times$ & \textbf{F}$\uparrow$ & \textbf{C}$\uparrow$ & \textbf{MR}$\downarrow$ & $\times$ & \textbf{F}$\uparrow$ & \textbf{C}$\uparrow$ & \textbf{MR}$\downarrow$ & $\times$ & \textbf{F}$\uparrow$ & \textbf{C}$\uparrow$ & \textbf{MR}$\downarrow$ & $\times$ & \textbf{F}$\uparrow$ & \textbf{C}$\uparrow$ & \textbf{MR}$\downarrow$ & $\times$ \\

\midrule
Texture & 1.00 & 0.88 & \textbf{1.57} & 1e0 & 1.00 & 0.88 & \textbf{1.42} & 1e0 & 1.00 & 0.91 & \textbf{1.14} & 1e0 & 1.00 & 0.88 & 6.00 & 2e0 & 1.00 & 0.97 & \textbf{1.71} & 1e0 & 1.00 & 0.90 & \textbf{1.92} & 2e0 \\
Dec.+Text. & 0.72 & 0.74 & 6.71 & 2e1 & 0.93 & 0.84 & 4.85 & 2e1 & 0.95 & 0.91 & 1.57 & 1e1 & 0.91 & 0.82 & \textbf{5.85} & 2e1 & 0.97 & 0.95 & 2.00 & 9e0 & 0.89 & 0.85 & 3.75 & 2e1 \\
\midrule
Struct.-100 & 0.82 & 0.93 & \textbf{1.71} & 3e3 & 0.91 & 0.90 & \textbf{1.14} & 5e2 & 0.85 & 0.74 & \textbf{2.28} & 7e2 & 0.84 & 0.89 & \textbf{3.57} & 1e3 & 0.92 & 0.89 & 3.42 & 5e2 & 0.85 & 0.88 & \textbf{2.51} & 2e3 \\
Struct.-250 & 0.83 & 0.89 & 3.00 & 5e3 & 0.86 & 0.85 & 2.71 & 3e2 & 0.84 & 0.91 & 2.42 & 8e2 & 0.81 & 0.86 & 4.57 & 1e3 & 0.93 & 0.90 & \textbf{3.00} & 6e2 & 0.85 & 0.89 & 3.18 & 2e3 \\
Struct.-500 & 0.87 & 0.91 & 2.57 & 1e4 & 0.84 & 0.87 & 4.71 & 3e2 & 0.84 & 0.86 & 5.14 & 8e2 & 0.89 & 0.89 & 5.00 & 1e3 & 0.87 & 0.90 & 4.71 & 6e2 & 0.85 & 0.89 & 4.24 & 3e3 \\
Struct.-750 & 0.84 & 0.84 & 3.14 & 1e4 & 0.82 & 0.79 & 6.42 & 4e2 & 0.81 & 0.83 & 6.42 & 8e2 & 0.89 & 0.88 & 5.00 & 1e3 & 0.85 & 0.89 & 3.28 & 7e2 & 0.83 & 0.86 & 5.12 & 3e3 \\
\midrule
Sem.-250 & 0.54 & 0.75 & 5.85 & 2e4 & 0.53 & 0.69 & \textbf{3.28} & 4e3 & 0.71 & 0.78 & \textbf{5.14} & 5e4 & 0.76 & 0.86 & \textbf{0.28} & 9e3 & 0.62 & 0.85 & \textbf{5.57} & 8e3 & 0.61 & 0.80 & \textbf{4.49} & 3e4 \\
Sem.-100  & 0.32 & 0.76 & \textbf{5.42} & 1e5 & 0.57 & 0.70 & 4.71 & 9e3 & 0.63 & 0.83 & \textbf{5.14} & 2e5 & 0.64 & 0.85 & 2.85 & 2e4 & 0.52 & 0.89 & 5.71 & 2e4 & 0.52 & 0.81 & 4.99 & 8e4 \\
Sem.-50 & 0.42 & 0.68 & 6.00 & 1e6 & 0.23 & 0.75 & 6.71 & 2e4 & 0.42 & 0.82 & 6.71 & 3e5 & 0.58 & 0.86 & 2.85 & 4e4 & 0.55 & 0.83 & 6.57 & 3e4 & 0.44 & 0.81 & 5.77 & 3e5 \\
\bottomrule
\end{tabular}
\caption{\textbf{Compression Evaluation}: The F-score (F) measures the structural correlation with the original object, the CLIP score (C) measures the semantic distance with the original, the mean ranking (MR) is the average human ranking, and $\times$ is the compression ratio. The final average column averages each score across the seven objects in Figure~\ref{fig:examples}. Bold values have the best scores in their section.}
\label{tab:ranking_methods}
\end{table*}

For perceptual compression techniques like semantic compression, accurate automated evaluation is difficult since by definition their quality is bound to human perception
This problem is shared with modern large language models, which are notoriously difficult to evaluate in a fair automated way.
In these cases, human evaluation is often preferred through public leaderboards, like the Chatbot Arena~\cite{chiang2024chatbotarenaopenplatform} and K-Sort Arena~\cite{li2024ksortarena}, in combination with automated task-specific evaluations.
We follow this approach and focus on human evaluation and augment it with two automated methods: F-score, for structural similarity, and CLIP for semantic similarity.

We evaluate these methods in Table~\ref{tab:ranking_methods} on the same objects in Figure~\ref{fig:examples}.
These objects were taken from Objaverse~\cite{deitke2022objaverse} and chosen because they contain enough internal details to show differences between methods and because they vary in complexity and file size, between 90KB for the warrior and 40MB for the cup.
Significantly more examples are shown in Appendix~\ref{sec:additional}.

\subsection{F-Score}
\label{sec:f_score}

The F-score measures the structural overlap between the original and the decompressed object meshes.
This score first randomly sub-samples the meshes to produce two sets of points: 
\( S_{\text{orig}} \) and \( S_{\text{decomp}} \).
Then, it calculates the relative overlap of these two sets following the expression:

   \[
   \frac{\left| \{ p \in S_{\text{1}} : \exists q \in S_{\text{2}}, \; \| p - q \| \leq d \} \right|}{|S_{\text{1}}|}
   \]
where the distance threshold $d$ is a parameter that indicates whether two points are close enough to be a match.
The threshold $d$ is set to 0.05 for all measurements in this section.
The precision $P$ and recall $R$ are calculated with this formula, where for precision \( S_1 = S_{\text{decomp}} \) and \( S_2 = S_{\text{orig}} \), and vice versa for recall.
These two quantities are averaged using the harmonic mean to form the F-score.

   \[
   F = \frac{2 \cdot \text{P} \text{R}}{\text{P} + \text{R}}
   \]

For traditional methods, higher F-scores indicate better compression since the original structure is preserved.
Yet, the usefulness of this score diminishes with semantic compression since at these compression rates, the exact structural overlap is not important.
For example, in Figure~\ref{fig:structure}, if the creature had more patches of leaves or hopped on its other foot, its quality would be approximately the same to most people. Therefore, this score is primarily useful for structural semantic compression. 

\subsection{CLIP}
\label{sec:clip}

In addition, we automatically evaluate the CLIP score~\cite{radford2021clip} between the frontal views of the original and decompressed objects.
Multiple models can be used with CLIP, and for this section, we use the ViT-B/32~\cite{dosovitskiy2021vit} base model for a balance of quality and performance.
It generates the embedding for both views and then computes the cosine similarity between them to produce the final score.
This score is common in evaluating generated 3D objects, yet it typically can only measure similarity between coarse-grained semantics.
This makes it useful as a high level filter for semantic content, but like the F-score it does not properly track human preferences and it can be sensitive to structural changes like color and shape.

\subsection{Human Evaluation}
\label{sec:human}

Given the limitations of the automated methods, we rely on averaged human evaluations to be the gold standard quality metric for the decompressed objects.
Since there are few methods to compare, a simple \textit{mean rank} score is chosen to rank the compression quality of different methods and hyper-parameters.
For a given object, a list of decompressed objects is given, each computed by a different method.
Then, the participant ranks (zero-indexed) these in order from best to worst subjective quality, following the general prompt: "In a virtual world, which of the compressed objects would you prefer?"
This prompt is broad to allow the participant to naturally balance accuracy to the original, aesthetics, and other characteristics.
These ranks are then averaged across participants to produce the final score.

\subsection{Discussion}
\label{sec:discussion}

Table~\ref{tab:ranking_methods} lists the evaluation results across the same set of objects chosen in Figure~\ref{fig:examples}.
It measures the F, CLIP, and mean rank scores for the traditional, structured semantic, and semantic methods.
As expected, pure texture compression does not affect the F-score, and semantic compression leads to large decreases since the object is often a different shape.
Also, the higher F values in the structural semantic section show that the 2D structural control leads to consistent 3D shapes.  

Yet, in general, these results hint at the weakness of the F and CLIP scores in evaluating semantic compression since they do not strongly track the MR scores.
Also, the MR score shows average improvements with larger file sizes, which matches the expected quality-compression tradeoff from Figure~\ref{fig:semantic}.
This means that the best performing semantic methods have the largest number of characters, and the best structured semantic methods usually have the most structural detail.
Altogether, semantic methods, with and without structural detail, often beat traditional compression when it is pushed to its limits.
\section{Related Work}
\label{sec:related}

Recent work has defined the field of semantic compression, first even through the use of direct human labor.
Bhown et al. used humans to directly interpret and describe images to another human who used image editing software to reconstruct the original~\cite{bhown2018humans}.
This work showed that human-guided, semantic compression led to human-rated improvements over existing techniques at large compression ratios, yet it was clearly not scalable given the human in the loop.
Later, Dotzel et al.~\cite{dotzel2024microbits} expanded on this concept and automated it using the strong multi-modal capabilities of OpenAI ChatGPT~\cite{openai2023gpt4}.
This demonstrated the promise of using multi-modal large language models as substitutes for humans in the semantic compression pipeline and attempted to map out the quality-performance curves for image compression, attaining results up to $1000\times$ smaller than JPEG by allowing flexibility in structural information.
In addition, Lei et al.~\cite{lei2023text} used conditional diffusion for image compression by extracting the edge maps from images and natural-language descriptions.
\section{Conclusion}
\label{sec:conclusion}

Semantic compression represents a paradigm shift compared to structural compression, and recent advancements within 3D generative models have enabled expanded its domain to include 3D objects.
This work demonstrates the feasibility of semantic compression for 3D objects, with a pipeline built from public generative models, achieving up to $10^5\times$ compression rates while preserving perceptual quality on some objects.
It also highlights the limitations with current technology, and the future potential of the next generation of generative models that learn even stronger representations over 3D objects.
Further combining structural and semantic compression leads to more practical compression around $100-1000\times$ ratios with more acceptable quality losses.
These challenges align with the needs of emerging platforms like the metaverse, where large collections of objects and collaborative environments benefit from the scalability and portability of semantic formats.
By addressing these limitations, future work can target these emerging applications to improve the quality-compression tradeoffs.

{
    \small
    \bibliographystyle{ieeenat_fullname}
    \bibliography{main}

\begin{thebibliography}{21}
\providecommand{\natexlab}[1]{#1}
\providecommand{\url}[1]{\texttt{#1}}
\expandafter\ifx\csname urlstyle\endcsname\relax
  \providecommand{\doi}[1]{doi: #1}\else
  \providecommand{\doi}{doi: \begingroup \urlstyle{rm}\Url}\fi

\bibitem[Bhown et~al.(2018)Bhown, Mukherjee, Yang, Chandak, Fischer-Hwang, Tatwawadi, and Weissman]{bhown2018humans}
Ashutosh Bhown, Soham Mukherjee, Sean~T. Yang, Shubham Chandak, Irena Fischer-Hwang, Kedar Tatwawadi, and Tsachy Weissman.
\newblock Humans are still the best lossy image compressors.
\newblock \emph{2019 Data Compression Conference (DCC)}, 2018.

\bibitem[Canny(1986)]{canny1986canny}
John Canny.
\newblock A computational approach to edge detection.
\newblock \emph{IEEE Transactions on Pattern Analysis and Machine Intelligence}, 1986.

\bibitem[Chiang et~al.(2024)Chiang, Zheng, Sheng, Angelopoulos, Li, Li, Zhang, Zhu, Jordan, Gonzalez, and Stoica]{chiang2024chatbotarenaopenplatform}
Wei-Lin Chiang, Lianmin Zheng, Ying Sheng, Anastasios~Nikolas Angelopoulos, Tianle Li, Dacheng Li, Hao Zhang, Banghua Zhu, Michael Jordan, Joseph~E. Gonzalez, and Ion Stoica.
\newblock Chatbot arena: An open platform for evaluating llms by human preference, 2024.

\bibitem[Deitke et~al.(2023{\natexlab{a}})Deitke, Liu, Wallingford, Ngo, Michel, Kusupati, Fan, Laforte, Voleti, Gadre, VanderBilt, Kembhavi, Vondrick, Gkioxari, Ehsani, Schmidt, and Farhadi]{deitke2023objaversexl}
Matt Deitke, Ruoshi Liu, Matthew Wallingford, Huong Ngo, Oscar Michel, Aditya Kusupati, Alan Fan, Christian Laforte, Vikram Voleti, Samir~Yitzhak Gadre, Eli VanderBilt, Aniruddha Kembhavi, Carl Vondrick, Georgia Gkioxari, Kiana Ehsani, Ludwig Schmidt, and Ali Farhadi.
\newblock Objaverse-xl: A universe of 10m+ 3d objects, 2023{\natexlab{a}}.

\bibitem[Deitke et~al.(2023{\natexlab{b}})Deitke, Schwenk, Salvador, Weihs, Michel, VanderBilt, Schmidt, Ehsani, Kembhavi, and Farhadi]{deitke2022objaverse}
Matt Deitke, Dustin Schwenk, Jordi Salvador, Luca Weihs, Oscar Michel, Eli VanderBilt, Ludwig Schmidt, Kiana Ehsani, Aniruddha Kembhavi, and Ali Farhadi.
\newblock Objaverse: A universe of annotated 3d objects.
\newblock \emph{CVPR}, 2023{\natexlab{b}}.

\bibitem[Dosovitskiy et~al.(2021)Dosovitskiy, Beyer, Kolesnikov, Weissenborn, Zhai, Unterthiner, Dehghani, Minderer, Heigold, Gelly, Uszkoreit, and Houlsby]{dosovitskiy2021vit}
Alexey Dosovitskiy, Lucas Beyer, Alexander Kolesnikov, Dirk Weissenborn, Xiaohua Zhai, Thomas Unterthiner, Mostafa Dehghani, Matthias Minderer, Georg Heigold, Sylvain Gelly, Jakob Uszkoreit, and Neil Houlsby.
\newblock An image is worth 16x16 words: Transformers for image recognition at scale.
\newblock \emph{ICLR}, 2021.

\bibitem[Dotzel et~al.(2024)Dotzel, Kotb, Dotzel, Abdelfattah, and Zhang]{dotzel2024microbits}
Jordan Dotzel, Bahaa Kotb, James Dotzel, Mohamed Abdelfattah, and Zhiru Zhang.
\newblock Exploring the limits of semantic image compression at micro-bits per pixel.
\newblock \emph{Tiny Papers Workshop at ICLR}, 2024.

\bibitem[Garland and Heckbert(1997)]{garland2023mesh}
Michael Garland and Paul~S. Heckbert.
\newblock Surface simplification using quadric error metrics.
\newblock \emph{SIGGRAPH}, 1997.

\bibitem[Gemini et~al.(2024)Gemini, Anil, Borgeaud, Alayrac, Yu, Soricut, Schalkwyk, Dai, Hauth, Millican, Silver, Johnson, Antonoglou, Schrittwieser, Ramachandruni, Zeng, Bariach, Weidinger, Vu, Andreev, He, Hui, Kashem, Subramanya, Hsiao, Hassabis, Kavukcuoglu, Sadovsky, Le, Strohman, Wu, Petrov, Dean, and Vinyals]{gemini2024gemini}
Team Gemini, Rohan Anil, Sebastian Borgeaud, Jean-Baptiste Alayrac, Jiahui Yu, Radu Soricut, Johan Schalkwyk, Andrew~M. Dai, Anja Hauth, Katie Millican, David Silver, Melvin Johnson, Ioannis Antonoglou, Julian Schrittwieser, Lakshmi Ramachandruni, Xiangkai Zeng, Ben Bariach, Laura Weidinger, Tu Vu, Alek Andreev, Antoine He, Kevin Hui, Sheleem Kashem, Amar Subramanya, Sissie Hsiao, Demis Hassabis, Koray Kavukcuoglu, Adam Sadovsky, Quoc Le, Trevor Strohman, Yonghui Wu, Slav Petrov, Jeffrey Dean, and Oriol Vinyals.
\newblock Gemini: A family of highly capable multimodal models.
\newblock \emph{arxiv}, 2024.

\bibitem[Groeneveld et~al.(2024)Groeneveld, Beltagy, Walsh, Bhagia, Kinney, Tafjord, Jha, Ivison, Magnusson, Wang, Arora, Atkinson, Authur, Chandu, Cohan, Dumas, Elazar, Gu, Hessel, Khot, Merrill, Morrison, Muennighoff, Naik, Nam, Peters, Pyatkin, Ravichander, Schwenk, Shah, Smith, Strubell, Subramani, Wortsman, Dasigi, Lambert, Richardson, Zettlemoyer, Dodge, Lo, Soldaini, Smith, and Hajishirzi]{groeneveld2024olmo}
Dirk Groeneveld, Iz Beltagy, Pete Walsh, Akshita Bhagia, Rodney Kinney, Oyvind Tafjord, Ananya~Harsh Jha, Hamish Ivison, Ian Magnusson, Yizhong Wang, Shane Arora, David Atkinson, Russell Authur, Khyathi~Raghavi Chandu, Arman Cohan, Jennifer Dumas, Yanai Elazar, Yuling Gu, Jack Hessel, Tushar Khot, William Merrill, Jacob Morrison, Niklas Muennighoff, Aakanksha Naik, Crystal Nam, Matthew~E. Peters, Valentina Pyatkin, Abhilasha Ravichander, Dustin Schwenk, Saurabh Shah, Will Smith, Emma Strubell, Nishant Subramani, Mitchell Wortsman, Pradeep Dasigi, Nathan Lambert, Kyle Richardson, Luke Zettlemoyer, Jesse Dodge, Kyle Lo, Luca Soldaini, Noah~A. Smith, and Hannaneh Hajishirzi.
\newblock Olmo: Accelerating the science of language models.
\newblock \emph{SIGGRAPH}, 2024.

\bibitem[Lei et~al.(2023)Lei, Uslu, Hassani, and Bidokhti]{lei2023text}
Eric Lei, Yiğit~Berkay Uslu, Hamed Hassani, and Shirin~Saeedi Bidokhti.
\newblock Text + sketch: Image compression at ultra low rates.
\newblock \emph{International Conference on Machine Learning Workshop on Neural Compression}, 2023.

\bibitem[Li et~al.(2024)Li, Liu, Fu, Li, Gu, Keutzer, and Dong]{li2024ksortarena}
Zhikai Li, Xuewen Liu, Dongrong Fu, Jianquan Li, Qingyi Gu, Kurt Keutzer, and Zhen Dong.
\newblock K-sort arena: Efficient and reliable benchmarking for generative models via k-wise human preferences, 2024.

\bibitem[Liu et~al.(2024)Liu, Xu, Jin, Chen, Varma~T, Xu, and Su]{liu2023one2345}
Minghua Liu, Chao Xu, Haian Jin, Linghao Chen, Mukund Varma~T, Zexiang Xu, and Hao Su.
\newblock One-2-3-45: Any single image to 3d mesh in 45 seconds without per-shape optimization.
\newblock \emph{Advances in Neural Information Processing Systems}, 36, 2024.

\bibitem[Long et~al.(2023)Long, Guo, Lin, Liu, Dou, Liu, Ma, Zhang, Habermann, Theobalt, and Wang]{Long2023Wonder3DSI}
Xiaoxiao Long, Yuanchen Guo, Cheng Lin, Yuan Liu, Zhiyang Dou, Lingjie Liu, Yuexin Ma, Song-Hai Zhang, Marc Habermann, Christian Theobalt, and Wenping Wang.
\newblock Wonder3d: Single image to 3d using cross-domain diffusion.
\newblock \emph{CVPR}, 2023.

\bibitem[OpenAI(2023)]{openai2023gpt4}
OpenAI.
\newblock Gpt-4 technical report.
\newblock \emph{arxiv}, 2023.

\bibitem[Qian et~al.(2025)Qian, Mai, Hamdi, Ren, Siarohin, Li, Lee, Skorokhodov, Wonka, Tulyakov, and Ghanem]{Qian2023Magic123OI}
Guocheng Qian, Jinjie Mai, Abdullah Hamdi, Jian Ren, Aliaksandr Siarohin, Bing Li, Hsin-Ying Lee, Ivan Skorokhodov, Peter Wonka, S. Tulyakov, and Bernard Ghanem.
\newblock Magic123: One image to high-quality 3d object generation using both 2d and 3d diffusion priors.
\newblock \emph{ICLR}, 2025.

\bibitem[Radford et~al.(2021)Radford, Kim, Hallacy, Ramesh, Goh, Agarwal, Sastry, Askell, Mishkin, Clark, Krueger, and Sutskever]{radford2021clip}
Alec Radford, Jong~Wook Kim, Chris Hallacy, Aditya Ramesh, Gabriel Goh, Sandhini Agarwal, Girish Sastry, Amanda Askell, Pamela Mishkin, Jack Clark, Gretchen Krueger, and Ilya Sutskever.
\newblock Learning transferable visual models from natural language supervision.
\newblock \emph{ICML}, 2021.

\bibitem[Rombach et~al.(2022)Rombach, Blattmann, Lorenz, Esser, and Ommer]{rombach2022stable}
Robin Rombach, Andreas Blattmann, Dominik Lorenz, Patrick Esser, and Björn Ommer.
\newblock High-resolution image synthesis with latent diffusion models.
\newblock \emph{CVPR}, 2022.

\bibitem[Shi et~al.(2023)Shi, Chen, Zhang, Liu, Xu, Wei, Chen, Zeng, and Su]{shi2023zero123plus}
Ruoxi Shi, Hansheng Chen, Zhuoyang Zhang, Minghua Liu, Chao Xu, Xinyue Wei, Linghao Chen, Chong Zeng, and Hao Su.
\newblock Zero123++: a single image to consistent multi-view diffusion base model, 2023.

\bibitem[Weissman(2023)]{weissman2023textual}
Tsachy Weissman.
\newblock Toward textual transform coding.
\newblock \emph{arxiv}, 2023.

\bibitem[Zhang et~al.(2023)Zhang, Rao, and Agrawala]{zhang2023adding}
Lvmin Zhang, Anyi Rao, and Maneesh Agrawala.
\newblock Adding conditional control to text-to-image diffusion models, 2023.

\end{thebibliography}
}

\clearpage
\setcounter{page}{1}
\maketitlesupplementary
\appendix

\section{Prompts}
\label{sec:prompts}

DESCRIBE = """Describe this object in as much detail as possible so that it's possible to recreate it based only on your description. Focus only on the object itself and not the background. Describe the shape of the object, its orientation, its colors, patterns, features, feature sizes in a single paragraph. Describe the main general single object and NOT some parts as multiple objects."""

SIMPLIFY =  """Reduce this description into the most important words. Only use lowercase letters. Restrict the response to this number of characters: """

\section{Edge Map Compression}
\label{sec:edge_map_compression}

By tuning this threshold, the edge map can focus only on significant contours and ignore the smaller detail in the image, which makes its overall structure relatively sparse.

In more detail, the returned edge map $I$ is a dense matrix in the form: $\{0, 1\}^{D \times D}$, the number of non-zeros is $N$, and the sparsity of the matrix can be represented with $S$ .
With sufficiently high $S$, sparse formats can reduce the size of the edge map compared to the dense format.
However, with lower $S$, the overhead from sparse format will be too large and lead to larger file sizes.

\begin{align*}
    & \textbf{I} \in \{0, 1\}^{D \times D}  \\
    & N = \sum_{i=1}^{D} \sum_{j=1}^{D} R_{i,j} \\
    & S = 1 - \frac{N}{D^2}
\end{align*}

First, the sparse coordinate format (COO) fits this case due the arbitrary shapes within edge maps of arbitrary images.
Other sparse formats assume additional structure such as denser rows, columns, or diagonals, which will not exist in natural images.
To examine whether a dense or sparse format should be used, we first calculate the break-even point.

For the COO format, we need \(log_2(D) \) bits per coordinate, and therefore with $B$ representing the total number of bits needed:
\[
B_{\text{dense}} = D^2
\]
\[
B_{\text{coord}} = N \times 2 \times \log_2(D)
\]

For coordinate format to be more space-efficient than dense format:
\[
B_{\text{coord}} < B_{\text{dense}}
\]

Solve for \( N \):
\[
N < \frac{D^2}{2 \times \log_2(D)}
\]

\[
\text{Density} = \frac{N}{D^2} < \frac{1}{2 \times \log_2(D)}
\]

The sparsity \( S \) is then:
\[
S = 1 - \text{Density} > 1 - \frac{1}{2 \times \log_2(D)}
\]

\[
S > 1 - \frac{1}{2 \times \log_2(D)}
\]

This formula provides the exact break-even point in terms of \( D \) for determining when coordinate format is more efficient than dense format.

For $2048\times2048$ images, which we use as part of the compression pipeline, the sparsity needs to be higher than 95\% for improvements for the sparse format.

\section{Examples}
\label{sec:examples}

\begin{figure*} [t]
    \centering
    \includegraphics[width=1\linewidth]{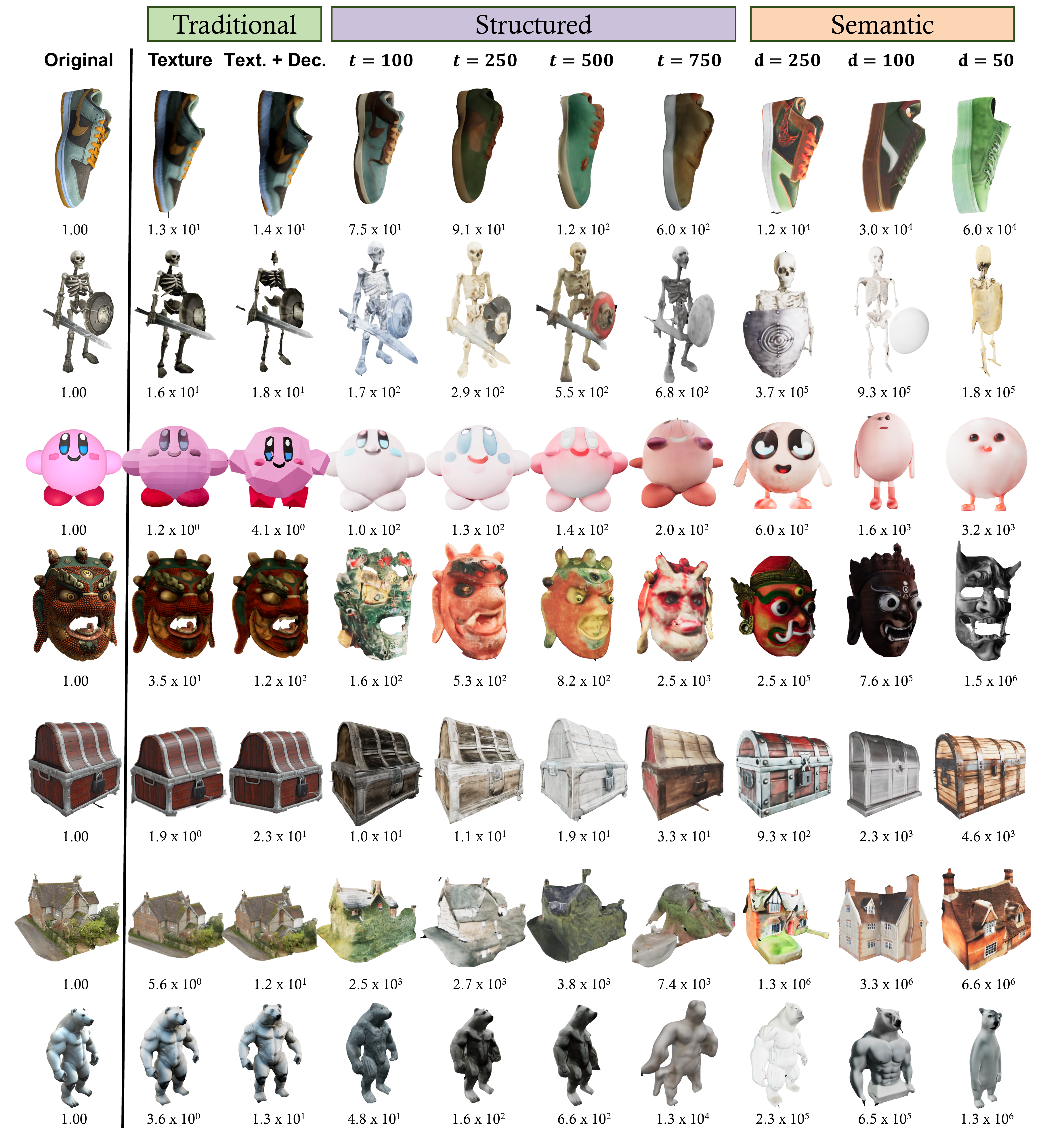}
    \vspace{-10pt}
    \caption{\textbf{Extra Compression Examples:} Traditional compression is limited to within $15\times$ for most objects while maintaining quality, while structured semantic and pure semantic methods can push orders of magnitudes further. The edge threshold $t$ controls the amount of detail in the edge map, and the character count $d$ limits the semantic content.}
    \label{fig:examples_extra}
\end{figure*}

Figure~\ref{fig:examples_extra} gives more examples across objects taken from Objaverse~\cite{deitke2022objaverse, deitke2023objaversexl} and from the online SketchFab repository.
It shows the tradeoffs across traditional, structured semantic and pure semantic compression.
Traditional compression very quickly loses its structure in its texture and mesh typically before hitting $10\times$ compression, although this varies significantly based on the original file size and balance between texture and mesh sizes.

The Kirby example (row 3), for example, is a very small file originally, around 90KB, and has little headroom for compression.
With just texture and mesh compression it already becomes warped with obvious polygons before $5\times$ compression.
Structured semantic compression naturally can maintain this structure better with a high-definition edge map around $100\times$ compression, where traditional methods cannot reach.
Then, pure semantic compression can push even further into $1000\times$ but it quickly becomes unrecognizable as Kirby as more and more features are lost, e.g., the size and color of the eyes and the existence of arms.
Likewise, the shoe example (row 1) shows the gradual loss of feature information with pure semantic compression.
It starts with the Nike swoop, then turns into a generic two-tone suede, and then finally a simpler pure green shoe.

Across examples the $d=250$ pure semantic category shows its power, matching the human ranking results in Table~\ref{tab:ranking_methods}.
Since these results are driven by the more powerful proprietary DALLE model, it hints at what better trained models with spatial conditioning can accomplish.
In most examples, even though the shape is different, this method produces a strong matching the main features of the original, e.g., show colors and swoop, skeleton with detailed shield, and pink creature with large eyes.
In the bear-human example (last row), it gives a very strong recreation of pose, musculature, and hybrid animal, even though it is clearly still different and not copied.

\section{Decimate}
\label{sec:decimate}

The decimate method is a common method for reducing the number of polygons within a 3D mesh by iteratively collapsing edges, vertices, or triangles.
This work uses the \textit{pyfqmr} library, which wraps an implementation of the quadratic edge collapse algorithm\footnote{\url{https://github.com/sp4cerat/Fast-Quadric-Mesh-Simplification}}.

This method only handles mesh decimation a nd ignores the texture information, which assumes the original topology.
Therefore, we update the uv-mapping between the texture and the mesh to allow remapping the textures after decimation.
This is further combined with texture JPEG compression to enable higher compression ratios.

\section{ControlNet}
\label{sec:controlnet}

ControlNet~\cite{zhang2023adding} is a popular work that builds on top of the open-source Stable Diffusion~\cite{rombach2022stable} to condition on spatial information.
It reuses the trained layers from Stable Diffusion and adds additional zero convolutions for conditioning control.
It uses multiple controls including edge, depth, segmentation, and human pose information, and in this work we only rely on the variant fine-tuned for edge maps.

\section{Additional Evaluation}
\label{sec:additional}

Table~\ref{tab:ranking_methods_additional} shows the evaluation for the remaining objects in Figure~\ref{fig:examples}.
These results show a similar trend to those in Table~\ref{tab:ranking_methods}, showing the effect of structural conditioning on the F score and strength of larger structured results determined by the human ranking.

\begin{table}[t]
\centering
\footnotesize
\setlength{\tabcolsep}{2pt}  
\begin{tabular}{lcccccccc}
\toprule
& \multicolumn{4}{c}{\textbf{Warrior}} & \multicolumn{4}{c}{\textbf{Cup}} \\
\cmidrule(lr){2-5} \cmidrule(lr){6-9}
\textbf{Method} & \textbf{F} & \textbf{C} & \textbf{HR} & $\times$ & \textbf{F} & \textbf{C} & \textbf{HR} & $\times$ \\
\midrule
Texture & 1.00 & \textbf{0.86} & 0.57 & 1.9e0 & 1.00 & \textbf{0.94} & 1.00 & 1.1e0 \\
Dec. + Text. & 0.89 & 0.72 & 1.85 & 4.5e0 & 0.88 & 0.95 & 3.42 & 4.5e1\\
\midrule
Structural-100 & 0.81 & 0.90 & \textbf{2.42} & 7.1e1 & 0.78 & 0.92 & 3.00 & 5.6e3 \\
Structural-250 & 0.83 & 0.94 & 4.14 & 7.3e1 & 0.83 & 0.84 & \textbf{2.42} & 7.1e3 \\
Structural-500 & 0.79 & 0.92 & 3.85 & 8.1e1 & 0.83 & 0.85 & 3.71 & 7.4e3 \\
Structural-750 & 0.78 & 0.92 & 4.85 & 8.3e1 & 0.81 & 0.86 & 6.71 & 7.6e3 \\
\midrule
Semantic-250 & 0.39 & 0.83 & \textbf{5.71}  & 4.3e2 & 0.69 & 0.86 & 5.57 & 1.2e5 \\
Semantic-100  & 0.42 & 0.86 & 5.71 & 1.3e3 & 0.52 & 0.79 & 5.42 & 2.5e5 \\
Semantic-50 & 0.48 & 0.85 & 6.85 & 2.5e3 & 0.41 & 0.86 & \textbf{4.71} & 5.4e5 \\
\bottomrule
\end{tabular}
\caption{\textbf{Compression Evaluation}: F represents the F score and measures the structural correlation with the original object, C is the CLIP score that measures the semantic distance with the original, and HR is the human ranking that was averaged over ten participants.}
\label{tab:ranking_methods_additional}
\end{table}

\end{document}